\documentclass[letterpaper, 10 pt, conference]{ieeeconf}  

\IEEEoverridecommandlockouts                              

\usepackage{graphics} 
\usepackage{epsfig} 
\usepackage{mathptmx} 
\usepackage{times} 
\usepackage{amsmath} 
\usepackage{amssymb}  
\usepackage{xcolor}
\usepackage{relsize}
\usepackage{hyperref}

\title{\LARGE \bf
Contrastively Learning Visual Attention as Affordance Cues from Demonstrations for Robotic Grasping
}

\author{Yantian Zha, Siddhant Bhambri and Lin Guan
\thanks{*This research is supported in part by ONR grants N00014-19-1-2119, N00014-16-1-2892, N00014-18-1-2442, N00014-18-1-2840, N00014-9-1-2119, AFOSR grant FA9550-18-1-0067, and a JP Morgan AI Faculty Research grant. The first author thanks his advisor Prof. Subbarao Kambhampati for his encouragement and support, Prof. Heni Ben Amor for a helpful discussion, and Allen Z. Ren for the communications with the first author that speeds up reproducing the baseline work.}
\thanks{All authors are with the School of CS \&  AI, Arizona State University, United States of America. Email:
        {\tt\small \{yzha3, sbhambr1, lguan9\}@asu.edu}}%
}

\begin{document}

\maketitle
\thispagestyle{empty}
\pagestyle{empty}

\begin{abstract}

Conventional works that learn grasping affordance from demonstrations need to explicitly predict grasping configurations, such as gripper approaching angles or grasping preshapes. Classic motion planners could then sample trajectories by using such predicted configurations. In this work, our goal is instead to fill the gap between affordance discovery and affordance-based policy learning by integrating the two objectives in an end-to-end imitation learning framework based on deep neural networks. From a psychological perspective, there is a close association between attention and affordance. Therefore, with an end-to-end neural network, we propose to learn affordance cues as visual attention that serves as a useful indicating signal of how a demonstrator accomplishes tasks, instead of explicitly modeling affordances. To achieve this, we propose a contrastive learning framework that consists of a Siamese encoder and a trajectory decoder. We further introduce a coupled triplet loss to encourage the discovered affordance cues to be more affordance-relevant. Our experimental results demonstrate that our model with the coupled triplet loss achieves the highest grasping success rate in a simulated robot environment. Our project website can be accessed at \footnote{https://sites.google.com/asu.edu/affordance-aware-imitation/project}.

\end{abstract}

\section{INTRODUCTION}

\begin{figure}[!t]
\centering
\includegraphics[width=1\linewidth]{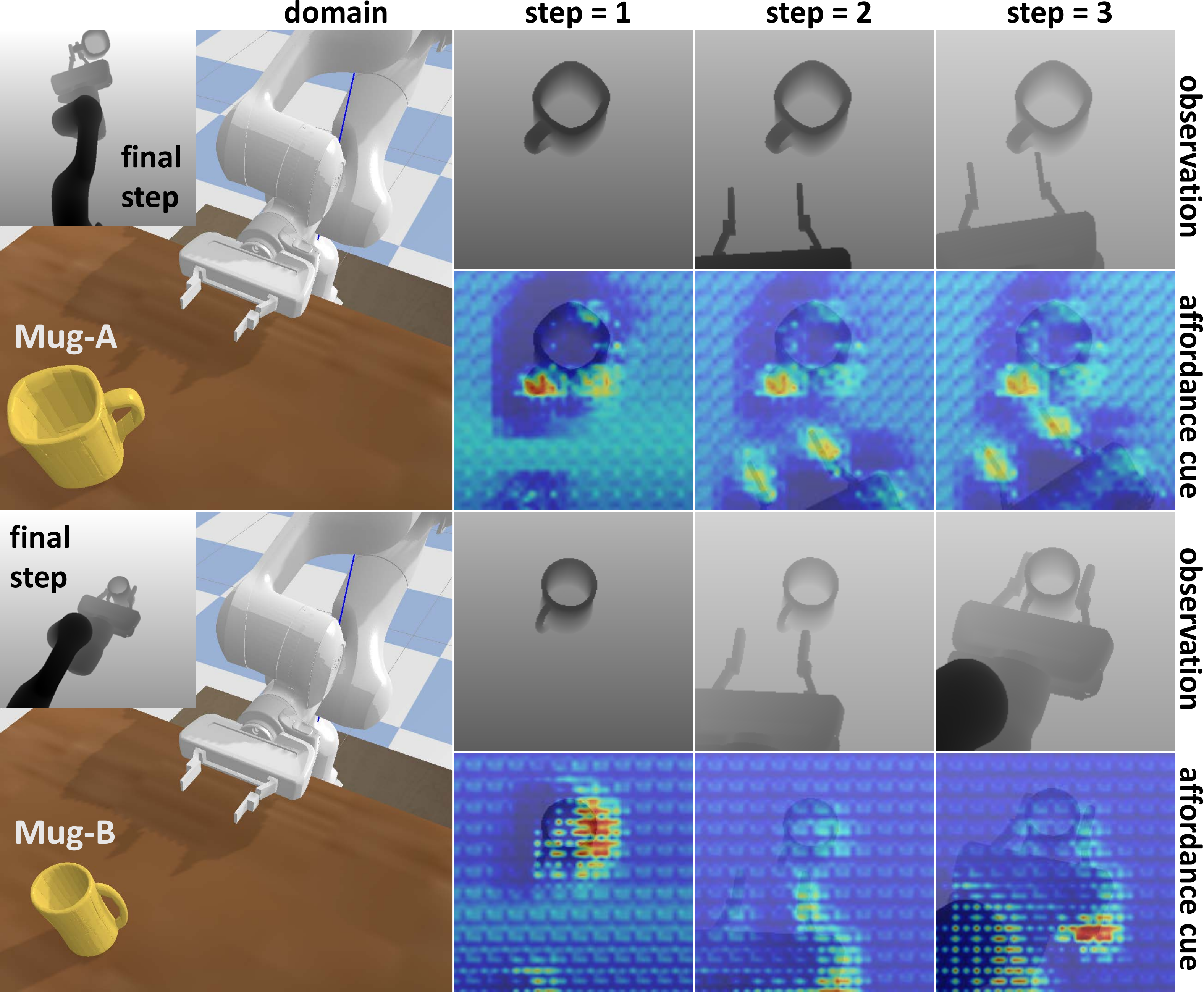} 
\caption{An example of two mugs: Humans should have different affordance-effect judgments on their bodies or handles. At each step, the robot takes one depth image (observation) and generates one attention map, a.k.a affordance cue, that further triggers an action. We also show how the robot eventually picks up a mug, either by holding the body or handle, by showing the final step observation.}
\label{fig:example}
\end{figure}

Humans tend to understand objects and their parts from potentially applicable actions or motion primitives that can achieve effects for accomplishing a task. This phenomenon is abstracted as an ecological psychology concept called affordance established by J. J. Gibson (\cite{gibson2014ecological}). An affordance defines a mapping from an object feature to all applicable actions (\cite{qian2020grasp}). Essentially, an affordance represents an object-action-effect relationship, which is an interactive procedure between an actor (e.g. a hand) and an object (e.g. a mug). Consider the example shown in Fig. \ref{fig:example}, in a mug grasping task, a human teacher’s affordance biases (affordance-effect judgments) might vary with the shape and size of the mug -- the mug-A is graspable from its handle, whereas the mug-B is graspable from its body. When a robot learns from human demonstrations, it would be beneficial if the robot also discovers such affordance bias behind human demonstrations and generate (visual) affordance cues to support its learning. 


There are several works (\cite{de2006learning,sweeney2007model,detry2013learning}) that propose to learn affordance knowledge from human demonstrations that avoid a labor-intensive process of collecting affordance labels. 
However, previous works fully decouple affordance discovery from behavior learning and execution, which means the affordance predictor and the robot controller are trained or constructed separately. In those methods, the predicted affordances are fed into classic motion planners in an engineered fashion. In this work, we instead combine the learning of affordance knowledge and motion generation from human demonstrations in an end-to-end deep imitation learning framework. We further argue that learning visual attention as affordance cues, rather than explicitly modeling affordances, is enough to be a reliable latent feature for the actor. The actor here is also a part of the whole neural network, instead of being a separate and unlearnable external planner. 

When learning affordance knowledge from demonstrations, visual attention can be an informative cue for inferring affordance. In fact, the close association between visual attention and affordance has been investigated by some cognitive psychology works (\cite{anderson2002attentional,kostov2012role}). According to them, humans use visual attention conditioned on objects' geometric and spatial properties to speed up affordance-effect judgments, which then helps generate motor signals (behaviors). 
When the attention comes from a controlled mental process (i.e. top-down attention) driven by a task, object parts that are permittable for accomplishing the task would be highlighted, from which proper affordance-effect judgments can be derived (\cite{riggio2008role}). Indeed, humans do not explicitly think about what are all possible ways of picking-up a mug at each of its parts or pixels once she or he already learned how to grasp different mugs. An earlier work (\cite{may2007gpu}) also investigates how visual attention could be associated with affordance cueing in the context of robotics.

In this work, we hypothesize that by encouraging the robot to attend to discriminative features that explain the differences between different demonstrated behaviors, the robot will be able to more effectively discover affordance information and better imitate human behavior. This is because when a human teaches affordance knowledge to a robot, the human tends to think about what makes he/she have different affordance-effect judgments and how could the resulting trajectories be accordingly different. Again taking Fig. \ref{fig:example} as an example, if the human shows two different trajectories for the robot to pick up the two different mugs, the robot could discover important discriminative features regarding mug shapes and sizes and consequently better learn from demonstrations. If the robot could attend to the handle of Mug-A, then it would be more likely for the robot to grasp the Mug-A by its handle, rather than its body.

Therefore, our central problem is to help robots to imitate humans not only at the behavior level but also with a hidden objective of discovering the affordance-relevant distinctiveness underlying human demonstrations. As such, robot could attend to appropriate parts of a specific mug that hints on the same affordance in human's mind and therefore helps trigger a similar behavior to humans'. 
To address the problem, we propose to use a deep Siamese encoder and trajectory decoder that are trained jointly with a contrastive loss and a behavior cloning loss in an end-to-end fashion. We also propose a coupled triplet loss to encourage the discovered discriminative features to be more affordance-relevant.

To thoroughly evaluate our work, we compare our model with a variant version that uses a normal triplet loss, a version without using Siamese network, and a baseline ConvNet-based behavior cloning model. We evaluate those models in terms of the grasping success rates and visualizations of predicted affordance cues. We empirically show that our model with the coupled triplet loss performs the best.

To the best of our knowledge, our work is the first to combine grasping affordance learning and imitation learning from expert demonstrations based on deep neural networks.

\section{RELATED WORKS}
\subsection{Affordance Learning for Grasping}
Regarding learning affordances for grasping, the majority of previous works use ground-truth affordance labels to learn affordances for grasping (\cite{mandikar2020dexterous,yenchen2020learning, zeng2018robotic, steil2004situated}). \cite{mandikar2020dexterous} use thermal maps to learn the graspable positions of several household objects. \cite{yenchen2020learning} employ transfer learning from pre-trained vision models to pixel-wise affordance prediction networks that help the robot generalize over novel objects. \cite{zeng2018robotic} also uses pixel-level affordances to identify multi-grasp possibilities for objects present in a cluttered area. 

There are also works that propose to learn affordances from demonstrations or via imitation learning: \cite{hsiao2006imitation,de2006learning,sweeney2007model,detry2013learning}. All of these works share similar frameworks of using classic unsupervised learning (e.g. clustering) to identify gripper control parameters that would be fed into a motion planner. In \cite{hsiao2006imitation}, the affordance is represented by contact points for grasping an object. They use a predefined tracking configuration to reduce the number of potential contact points from demonstrations. After detecting a set of contact points on a new object, nearest-neighbor classification is used to identify a template grasp that matches their demonstration data on similar objects. In \cite{de2006learning,sweeney2007model,detry2013learning}, they define affordable actions in affordances as approaching angles (\cite{de2006learning}), grasp preshape hypotheses (\cite{sweeney2007model}), or grasping prototypes (\cite{detry2013learning}). Then they use clustering to find condensed representations of affordances. In this work, we leverage the expressiveness of deep neural networks to implicitly learn affordance knowledge from human demonstrations. By learning from demonstrations with a deep contrastive learning framework, we evade the need of using ground truth affordances. 

\subsection{Attention Guided Imitation Learning}
Imitation learning or learning from demonstrations \cite{morton2017simultaneous,ren2020generalization} has been at the core of teaching robots to perform object-manipulation tasks in a similar way to humans performing the same task. By combining visual attention with imitation learning, robots could learn the information better by focusing on smaller but more important regions in manipulation tasks. In \cite{abolghasemi2019pay}, authors use natural language descriptors that are specific to the task at hand to generate guided attention cues. The masked attention method places attention on the entire object that is to be grasped but does not focus on the specific region that the robot needs to interact with. A similar issue can be observed in \cite{rama2020attentive} where the model first captures attention features of the object by generating attention maps for different stages of the object manipulation task. But the robot, in this case, can only learn to imitate the task and can not decide how to perform a specific task in a variable environment setting. 
Hence, one potential advantage of our approach is that it allows the robot to learn the specific 
graspable points in scenarios where the shape of the object (mug and its handle, in this
case) can also vary restricting the possible graspable areas even for a human.

\subsection{Discriminative Feature Learning}
The objective of discriminative feature learning is to make sure that the learned features of deep neural networks can represent different inputs contrastively enough \cite{wen2016discriminative}. Usually, such learned features can be easily separated by k-nearest neighbors algorithms \cite{fukunaga1975branch}. Various approaches have been proposed to address the discriminative learning problem for deep neural networks. One popular approach is Siamese neural network \cite{bromley1994signature} that was proposed in 1994 for verifying signatures. Quite a few works used Siamese neural networks as a backbone for new deep network models of discriminative feature learning. Siamese neural network can be combined with ConvNets and trained with a Binary Cross Entropy loss as in \cite{koch2015siamese}, or triplet loss as in \cite{schroff2015facenet}). Recently deep learning based Siamese neural networks are also applied in many new applications, like face recognition (\cite{schroff2015facenet,song2019occlusion}), object discovery (\cite{srivastava2018large,huang2016object}), object co-segmentation (\cite{banerjee2019cosegnet,lu2019see,mukherjee2018object}), and re-identification as in \cite{zhang2017attributes,Nepovinnykh_2020_WACV}. 

Besides using Siamese neural networks for discriminative feature learning, \cite{wen2016discriminative} proposes to replace normal classification loss functions in ConvNets with the Center loss. Center loss works by minimizing the intra-class variations and meanwhile keeping the inter-class feature variations separable enough. The work \cite{yan2014prototype} proposes a prototype-based
discriminative feature learning (PDFL) method. The work \cite{li2018domain} follows a similar idea to \cite{wen2016discriminative} and designs a discriminative feature learning algorithm for domain adaption. 

\section{PROBLEM STATEMENT} \label{sec:ps}

Our goal is to encourage robots to learn from expert demonstrations more efficiently by taking advantage of discovering affordance-relevant distinctiveness underlying all demonstrations. To capture such distinctiveness, we could learn an attention model that predicts \textbf{affordance cues} from observations. We assume that humans have hidden affordance knowledge that is associated with visual cues. When focusing on a region of an object, humans also know what could be affordable grasps on that region. \textbf{The highlighting of such a region could serve as an affordance cue that supports the imitation learning of the learner itself.} 

\begin{figure}[!h]
\centering
\includegraphics[width=0.94\linewidth]{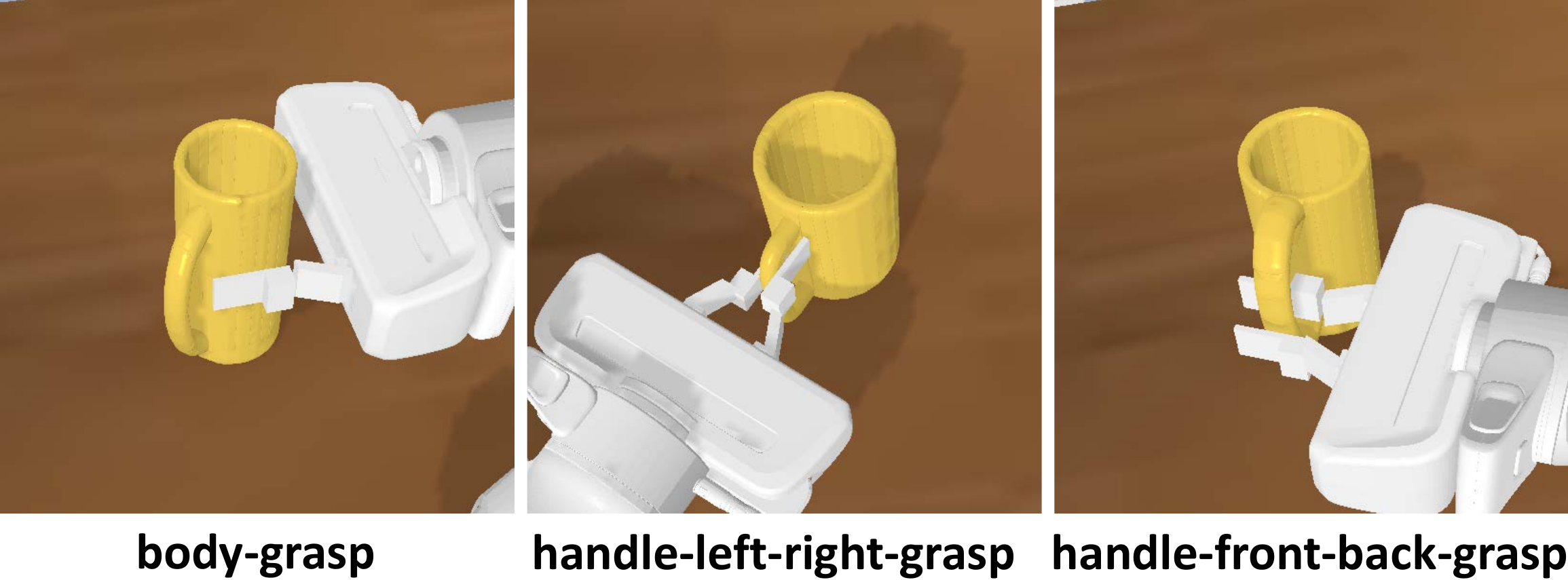}
\caption{The picture depicts three examples of the three candidate affordable grasps: body-grasp, handle-left-right-grasp, and handle-front-back-grasp.}
\label{fig:grasp_mp_exp}
\end{figure}

Our task is to pick up mugs in a way that allows pouring water in the near future. Hence, a robot could grasp a mug by reaching its gripper horizontally to the mug body, grasp the left and right sides of a handle, or grasp the front and back sides of a handle. Therefore, we have three candidate affordances (an object part and an applicable grasp): body-grasp, handle-left-right-grasp, and handle-front-back-grasp, as shown in Fig. \ref{fig:grasp_mp_exp}. 

We assume that robots share the same embodiment with humans: a human expert teleporates a robot to collect demonstration trajectories. The collected trajectories are categorized in terms of the three candidate affordances. Each trajectory $\tau$ is a sequence of triplets $(o_t,s_t,a^*_t)$: $\tau := \{ (o_t,s_t,a^*_t) \}_{t=1}^{T-1} \bigcup (s_T, o^*_T)$. $T$ denotes the trajectory length. $o_t$ denotes a depth image at step $t$. $s_t$ denotes a state vector at $t$ which has eight values: the relative 3-D position of the gripper to a mug, the relative 3-D Euler orientation of the gripper to mug, and two finger position values. $a^*_t$ denotes an expert action vector that has seven values: the translation of x, y, and z; the rotation of roll, pitch, and yaw; and a value that indicates if the fingers should be closed or not. All trajectories are categorized into $C$ \textbf{affordance categories} ($C$ is three in this work). In each category $c$, there are $N_c$ expert trajectories $\{ \tau_i^c \}_{i=1}^{N_c}$.

\section{APPROACH}

Our framework learns to reproduce humans' behavior with visual cues that hint at different affordances from human demonstrations. The learning of such visual cues is achieved by training a Siamese encoder, and the policy imitation is by a behavior-cloning-based trajectory decoder. The Siamese encoder and trajectory decoder are trained simultaneously in a contrastive learning framework.

\subsection{A Challenge and A Graphical Model}

Since human embeds her/his hidden knowledge of affordances into trajectories of different affordance categories, intuitively we could use contrastive learning to help discover affordance cues. A well-known contrastive loss is the triplet loss (\cite{schroff2015facenet}) defined in Equation  \ref{eq:trip_loss}.

\begin{align}
\begin{split} \label{eq:trip_loss}
    L(\mathcal{A}, \mathcal{P}, \mathcal{N}) = \\ \sum_{i=1}^N [\left \| f(\mathcal{A}_i)-f(\mathcal{P}_i) \right \|^2_2-\left \| f(\mathcal{A}_i)-f(\mathcal{N}_i) \right \|^2_2+M]_+
\end{split} 
\end{align} where $\mathcal{A}$, $\mathcal{P}$, and $\mathcal{N}$ denote the sets of anchor, positive, and negative trajectories; $\mathcal{A}_i$ denotes the $i$-th trajectory in $\mathcal{A}$ (likewise for $\mathcal{P}_i$ and $\mathcal{N}_i$); The $i$-th positive trajectory is sampled from the same affordance category of the $i$-th anchor trajectory, whereas the $i$-th negative trajectory is sampled from a different category. $M$ denotes a margin value, $\left \| z\right \|^2_2$ denotes a squared Euclidean distance metric, $[z]_+$ denotes any $z$ that is larger than zero, and, $f()$ is an encoding function that can be parameterized by a neural network.
 
Contrastive losses encourage a learner to discover recurring patterns in one category and discriminative patterns across different categories. From humans' perspective, shifting affordance cues would lead to the change of affordance-effect judgment such that a different action would be taken. Thus, for two trajectories that come from the same affordance category, their affordance cues would share certain similarities; for two trajectories that are sampled from different affordance categories, their affordance cues tend to be different. 


However, one potential challenge in using traditional contrastive learning frameworks is that the agent might learn to exploit affordance-irrelevant information (e.g. contexts, initial configurations) to distinguish two trajectories. Inspired by earlier affordance learning from demonstration works which extract grasping preshapes (\cite{detry2013learning}) or gripper-object approaching orientations (\cite{de2006learning}) to learn affordance clusters, we also extract \textbf{the segment of interaction state-action pairs} (shortly \textbf{interaction segment}) of each trajectory as an affordance-relevant feature. An interaction segment of a trajectory includes the state-action transitions that an actor (e.g. a gripper) changes the state of a target object. Since affordance is about object-action-effect relationships (\cite{lopes2007affordance}), we use the current state and previous expert action as the state-action features per step. Specifically speaking, we extract the interaction state-action transitions from the trajectory $\tau$:  $\{(s_t,a^*_{t-1})\}_{t=m}^n|_\tau$. The curly brackets $\{\}_{t=m}^n$ mean that the state of an object changes between the step $m$ and $n$ due to the motion of an actor. 

\begin{figure}[!h]
\centering
\includegraphics[width=0.8\linewidth]{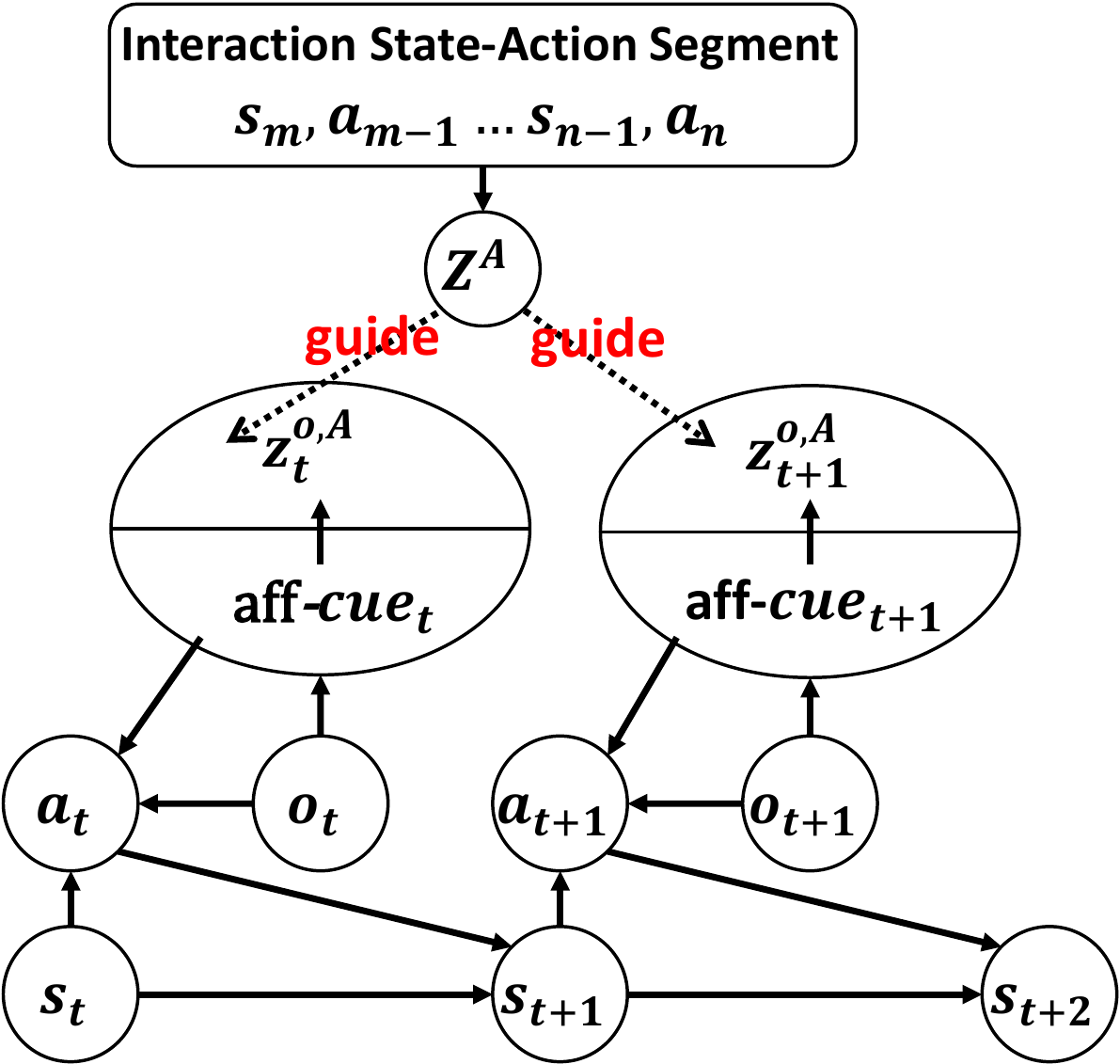} 
\caption{The graphical model of Siamese encoder. Each $\bigcirc$ denotes a node that represents a feature. Each directed edge from node $X$ to node $Y$ represents that ``Y'' depends on ``X''. The \text{\Large{$\ominus$}} denotes two nodes (features) that are generated by the same component one by one. The dashed edge means that the learning of affordance embedding $Z_A$ guides the learning of observation embedding $z_t^{o,A}$ at each step $t$.}
\label{fig:graphical_model_cs_net}
\end{figure}

\begin{figure*}[!h]
\centering
\includegraphics[width=16.5cm]{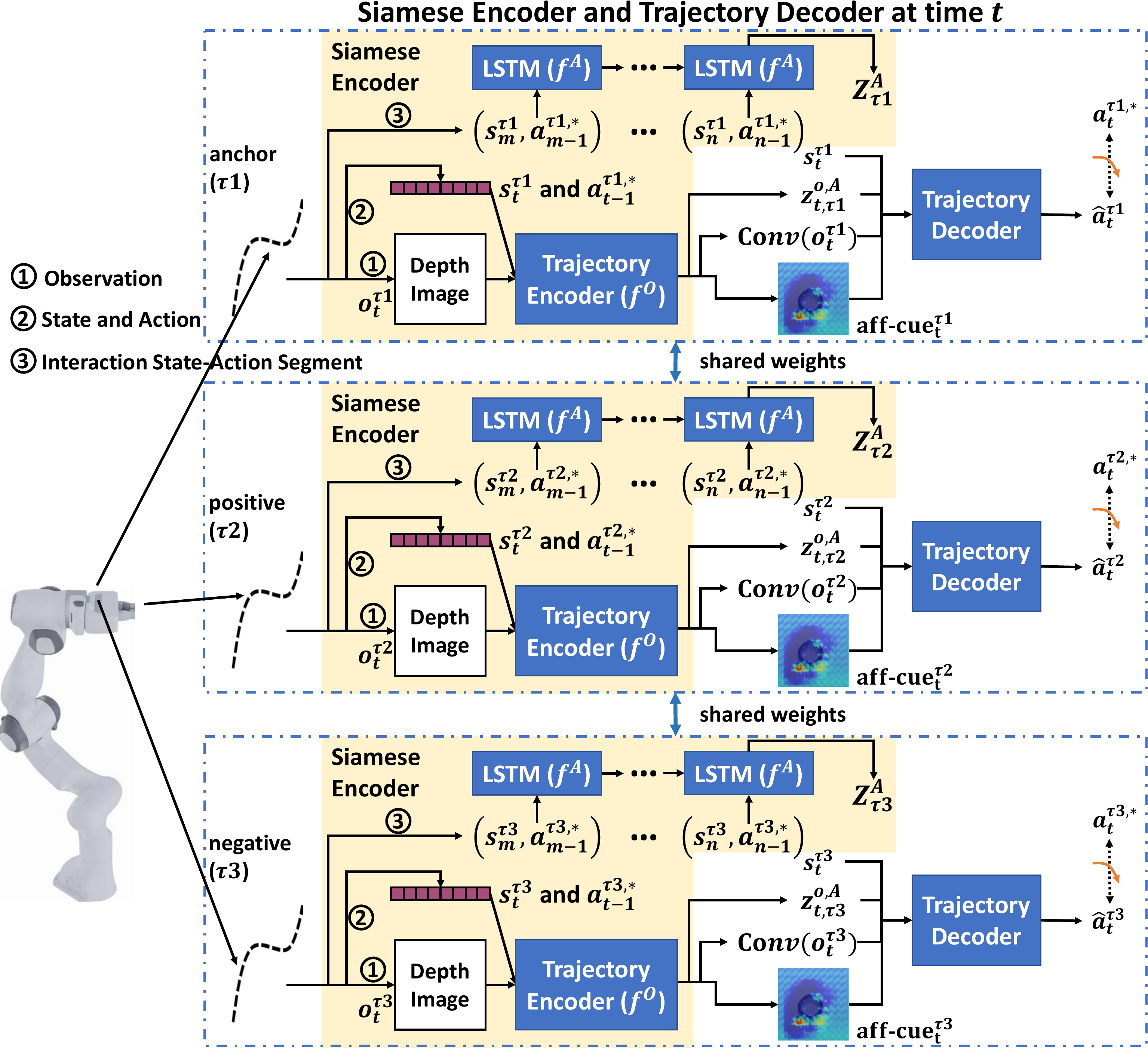} 
\put(-362,334){$\pmb{\mathcal{A}_i}$}
\put(-365,191){$\pmb{\mathcal{P}_i}$}
\put(-364,46){$\pmb{\mathcal{N}_i}$}
\caption{The overall contrastive learning architecture of Siamese encoder and trajectory decoder. The dashed edge with an orange arrow means that the distance between its connected features needs to be minimized.}
\label{fig:all_arch}
\end{figure*}



By contrastively learning from such interaction segments, we could learn a clean representation of affordances in the form of embeddings. Such embeddings could be named \textbf{affordance embeddings $Z^A$} which is computed by using Equ. \ref{eq:e1}. We do not directly condition robots' decision-making on $Z^A$. Instead, we treat $Z^A$ as a guidance to help robots learn an attention model that extracts an affordance-cue from an observation for decision-making. To achieve this objective, we feed state $s_t$, image $o_t$, and action $a_{t-1}^*$ at an arbitrary step $t$ of a trajectory into a neural network that generates latent features: an affordance-cue and a processed visual feature. We then convert such latent features to \textbf{observation embedding $z_t^{o,A}$} as illustrated in Equ. \ref{eq:e2}. In the rest of this paper, we use $Z^A$ and $Z^A_{\tau}$ to denote the same thing: an affordance embedding for an arbitrary trajectory ($\tau$). We use $Z^A_{\tau}$ when we need distinguish among different trajectories. Likewise for $z_t^{o,A}$ and $z_{t,\tau}^{o,A}$. Now we can encourage $z_t^{o,A}$ to be either closer or farther from $Z^A$ depending on if the trajectory that gives $z_t^{o,A}$ belongs to the same affordance category of the trajectory that gives $Z^A$ or not. This is essentially using the spatial relationships among embeddings in the space of $Z^A$ to guide the learning of observation encoding (Equ. \ref{eq:e2}). This way, we link the visual processing of high-dimensional sensory at any step of a trajectory to the affordance knowledge that can be learned faster from the lower-dimensional data of interaction segments.

\begin{gather}
Z^A_{\tau} = f^A(interaction\text{-}segment_\tau) = f^A(\{(s_t,a^*_{t-1})\}_{t=m}^n|_\tau) \label{eq:e1} \\
z_{t,\tau}^{o,A} = f^O(s_t^\tau,o^\tau_t,a^{\tau,*}_{t-1}) \label{eq:e2}
\end{gather} where $f^A()$ and $f^O()$ are two encoding functions whose neural network architectures are explained in Sec. \ref{sec:se}; $Z^A_{\tau}$ encodes the interaction segment of the trajectory $\tau$; and $z_{t,\tau}^{o,A}$ mainly encodes the high-dimension observation $o_t$ (e.g. a depth image) at the current step $t$ of the trajectory $\tau$ with auxiliary information like the current state $s_t$ and previous ground-truth action $a^*_{t-1}$.

The graphical model that describes the above idea is illustrated in Fig. \ref{fig:graphical_model_cs_net} which shows two transitions in a trajectory. We start by extracting the interaction state-action segment of this trajectory. The interaction segment is converted to the embedding $Z^A$ which guides the learning of $z_t^{o,A}$ as explained before. The current state $s_t$, observation $o_t$, and observation embedding $z_t^{o,A}$ determine what is the proper action $a_t$ to take. The generation of $z_t^{o,A}$ depends on an affordance-cue (aff-$cue_t$) that is extracted from $o_t$. Note that the embedding $z_t^{o,A}$ can essentially be viewed as a non-visual part of affordance-cue. However, in this paper, we focus on the visual part and we refer to the affordance-cue as an attention map.


\begin{figure*}[!h]
\centering
\includegraphics[width=17cm]{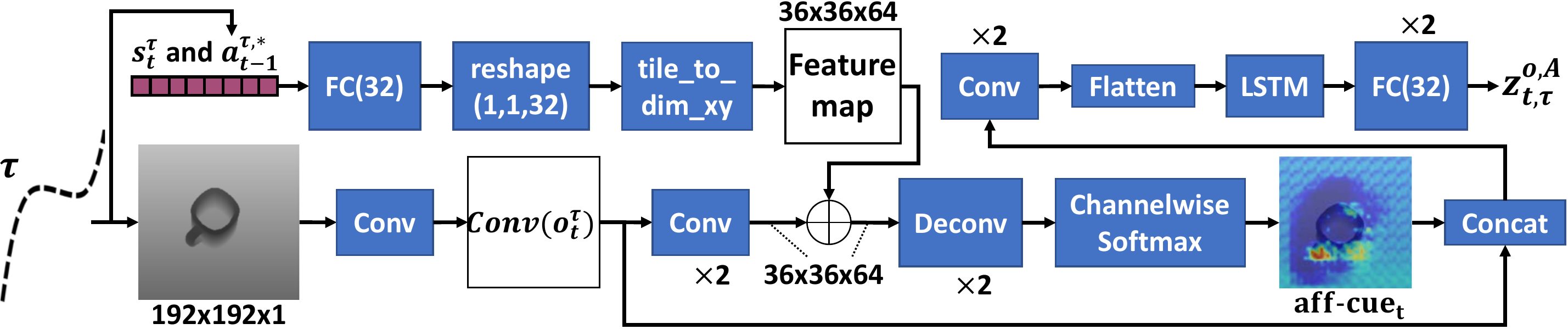}
\caption{The detailed architecture of the trajectory encoder module. ``$\times\#$'' denotes that a neural network component needs to be replicated $\#$ times.}
\label{fig:sim_encoder}
\end{figure*}

\subsection{Siamese Encoder and Coupled Triplet Loss}



Based on the graphical model, our design of Siamese encoder is depicted in the yellow region of Fig. \ref{fig:all_arch}. The Siamese encoder includes a LSTM layer to encode an interaction segment to generate $Z^A_\tau$ ($Z^A$ for a trajectory $\tau$). It also includes a trajectory encoder that encodes all information (image, state, and previous action) per step to generate $z^{o,A}_{t,\tau}$ ($z_t^{o,A}$ at step $t$ in a trajectory $\tau$). Fig. \ref{fig:all_arch} depicts the architecture at step $t$, but in the training phase, we feed into the Siamese encoder a trajectory of $T$ steps and would obtain a sequence of observation embeddings $\{z^{o,A}_{t,\tau}\}_{t=1}^T$. We also replicate a Siamese encoder into three copies and they share the same weights at any time during training. The three copies take three trajectories as inputs during training: an anchor, positive, and negative trajectory. The anchor and positive trajectories are sampled from the same affordance category of data, while the anchor and negative trajectories come from two different categories. Given a trajectory, each copy of Siamese encoder generates $Z^A_\tau$ and $\{z^{o,A}_{t,\tau}\}_{t=1}^T$ for different possible $\tau$ as explained before. Based on Equ. \ref{eq:trip_loss}, \ref{eq:e1} and \ref{eq:e2}, we propose the coupled triplet loss (Equ. \ref{eq:c_trip_loss}) to couple the learning of the two types of embeddings together.




\begin{gather}
\begin{split} \label{eq:c_trip_loss}
    L(\mathcal{A}, \mathcal{P}, \mathcal{N}) = \sum_{i=1}^N \{[\left \| Z^A_{\mathcal{A}_i}-Z^A_{\mathcal{P}_i}\right \|^2_2-\left \| Z^A_{\mathcal{A}_i}-Z^A_{\mathcal{N}_i}\right \|^2_2+M]_+ \\ + \sum_{t=1}^T[\left \| Z^{o,A}_{\mathcal{A}_i}-z^A_{t,\mathcal{P}_i}\right \|^2_2-\left \| Z^{o,A}_{\mathcal{A}_i}-z^A_{t,\mathcal{N}_i}\right \|^2_2+M]_+\} 
\end{split}
\end{gather} where $\mathcal{A}$, $\mathcal{P}$, and $\mathcal{N}$ are the anchor, positive, and negative sets of demonstration trajectories; $\mathcal{A}_i$ denotes the $i$-th trajectory in $\mathcal{A}$ (likewise for $\mathcal{P}_i$ and $\mathcal{N}_i$); $T$ denotes the length of an arbitrary trajectory; $f^A()$ and $f^O()$ are explained under the Equ. \ref{eq:e2}, which generates an affordance embedding $Z^A$ and a observation embedding $z_t^{o,A}$ respectively. $Z^A_{\tau}$ ($\tau$ could be either $\mathcal{A}_i$, $\mathcal{P}_i$, or $\mathcal{N}_i$) denotes an affordance embedding $Z^A$ for a trajectory $\tau$; Likewise, $z^{o,A}_{t,\tau}$ denotes an observation embedding $z^{o,A}_t$ for a trajectory $\tau$; the Sec. \ref{sec:se} explains $f^O()$ in more details. 

The coupled triplet loss can be decomposed into two contrastive learning objectives in the two square brackets $[]_+$ that are summed together. The first objective contrastively learns affordance embedding $Z^A$ from all extracted interaction segments. $Z^A$ could also be learned faster due to the low-dimensionality of interaction segment data. The second objective uses $Z^A$ to guide the learning of the observation embedding $z_t^{o,A}$. Note that $Z^{o,A}_{\mathcal{A}_i}$ is adjusted by both of the affordance embeddings $z^A_{t,\mathcal{P}_i}$ and $z^A_{t,\mathcal{N}_i}$. This way of formulating the coupled triplet loss provides a strong training signal for the observation encoder that generates $z_t^{o,A}$. In this sense, the learning of the embeddings $Z^A$ and $z_t^{o,A}$ are coupled together.





\subsection{The Details of Siamese Encoder} \label{sec:se}

The architecture of a Siamese encoder at an arbitrary time step is illustrated in the yellow region of Fig. \ref{fig:all_arch} and with more details in Fig. \ref{fig:sim_encoder}. The input can be either an anchor, positive, or negative trajectory. The definition of a trajectory is explained in Sec. \ref{sec:ps}. 

We start by explaining the detailed formulation of the encoding function $f^A()$. At the initial step of an input trajectory, we append a dummy action vector of all zeros to provide a dummy previous action for the first step. We extract the interaction segment between the step $m$ and $n$ and feed them into an LSTM network ($f^A()$) to generate the affordance encoding $Z^A$. In our work, the values of $m$ and $n$ are provided by a human expert. But in reality, $m$ and $n$ can be determined by tracking when the relative pose of the target object to gripper starts and stops changing.

We now explain the formulation of the encoding function $f^O()$. To process the observation encoding $Z^{o,A}_t$ per step $t$ of a whole trajectory, we feed it step-by-step into the trajectory encoder ($f^O()$) module of Siamese encoder. As depicted in Fig. \ref{fig:sim_encoder}, the trajectory encoder module is a two-branch architecture. The bottom branch is used to process visual features and the top branch is used to process low-dimensional features like state and action vectors. The top branch concatenates state and action together and feeds them to a fully-connected layer. After that, it tiles (repeats) the fully-connected layer feature along the x and y dimensions (e.g. $36\times 36$) of the visual feature map generated after the third convolutional layer at the bottom branch. This way, the feature map representation of low-dimension inputs is merged with the convolutional features of visual inputs by element-wise addition. Then two deconvolution layers are used to generate a two-channel feature map. After applying the feature map with a channel-wise Softmax, its two channels represent graspable and non-graspable probabilities per pixel respectively. We then extract the first channel of this feature map as an attention map (affordance-cue), aff-$cue_t$. Note that in our work, such attention maps are essentially latent attention maps because their spatial dimension matches that of the first convolution layer ($Conv(o_t^\tau)$) output, rather than the dimension of a raw input image. We then concatenate aff-$cue_t$ with $Conv(o_t^\tau)$. After this, the two convolutional layers, one LSTM network, and two fully connected layers are used to generate $Z^{o,A}_{t}$. In our work we set the encoding size of $Z^{o,A}_{t}$ to 32.

\begin{figure}[!h]
\centering
\includegraphics[width=1\linewidth]{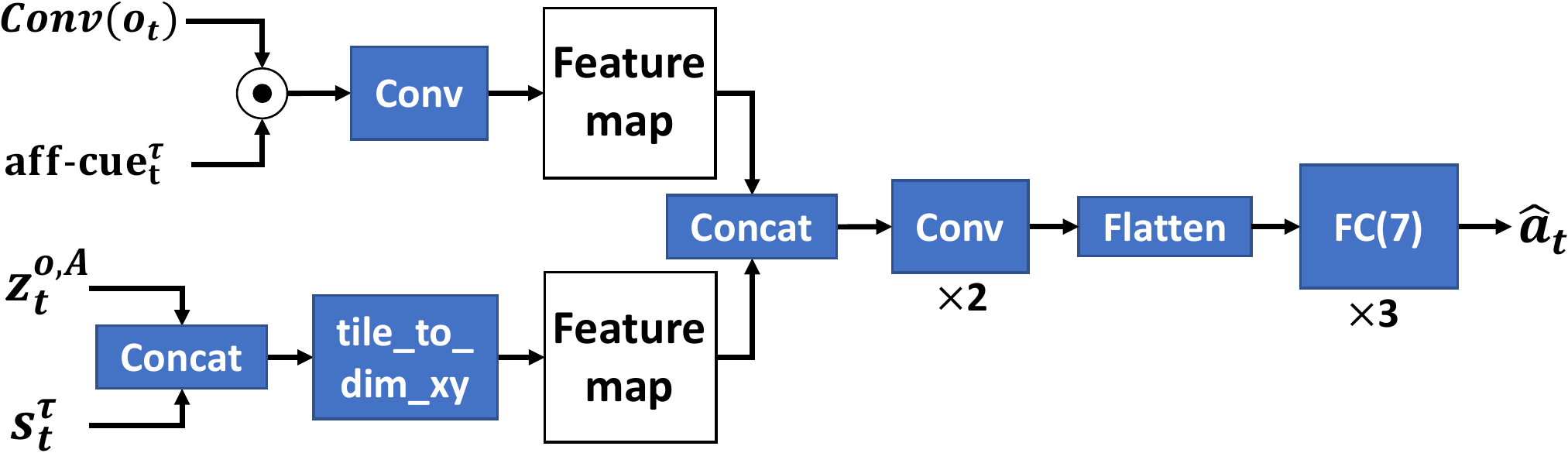}
\caption{The detailed architecture of the trajectory decoder. ``$\times\#$'' denotes that a neural network component needs to be replicated $\#$ times.}
\label{fig:traj_decoder}
\end{figure}

\subsection{Policy Network as Trajectory Decoder}
The design of our trajectory decoder is based on a convolutional policy network, as illustrated in Fig. \ref{fig:traj_decoder}. The inputs include current state $s_t$, the latent convolution feature $Conv(o_t)$, the affordance cue (aff-$cue_t$), and the contrastive embedding $z^{o,A}_t$. Since we treat a discovered affordance cue as an attention feature, we multiply aff-$cue_t$ with each channel of the convolution feature $Conv(o_t)$. 

We concatenate $s_t$ and $z^{o,A}_t$ together and obtain a new 1-D feature. We then tile (repeat) it across x and y dimensions of the visual feature map generated by the first convolutional layer at top branch. This way, we could concatenate the visual feature $Conv(o_t)$, state feature $s_t$, and contrastive embedding $z^{o,A}_t$ together along the channel dimension. This new feature is then fed into two convolution layers and three fully connected layers to obtain a predicted action $\hat{a}_t$.

The loss function for behavior decoding is based on a behavior cloning loss in Equ. \ref{eq:bc}. The overall loss function for training the entire Siamese encoder and trajectory decoder is a sum of the coupled triplet loss (Equ. \ref{eq:c_trip_loss}) and behavior cloning loss (Equ. \ref{eq:bc}), which enables a simultaneous learning of affordance knowledge and affordance-aware grasping from expert demonstrations.

\begin{gather}
loss_{bc} = \sum_{i=1}^N \{\sum_{t=1}^T [L_1(a^*_t, \hat{a}_t) + L_2(a^*_t, \hat{a}_t)]|_{\tau_i}\}  \label{eq:bc}
\end{gather} where $L_1$ and $L_2$ denotes L1-norm and L2-norm respectively; $T$ denotes the length of a trajectory; $|_{\tau_i}$ denotes that the ground-truth action $a^*_t$ is from the $i$-th trajectory $\tau_i$ in dataset.

\subsection{Testing a Trained Model}
\begin{figure}[!h]
\centering
\includegraphics[width=1\linewidth]{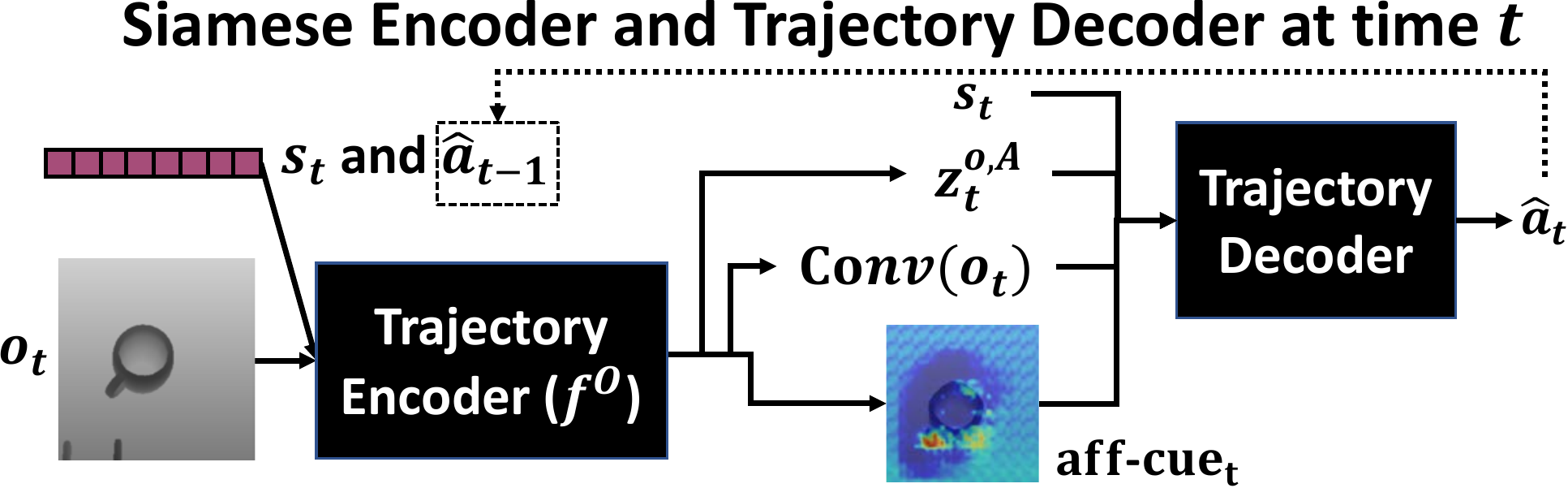}
\caption{The overall architecture of Siamese encoder and trajectory decoder in testing phase. The rectangular filled with black color means that the model weights are fixed. The dashed arrow and box represent that the prediction of action $\hat{a}_t$ at each step would be fed into the Trajectory Encoder at the next step.}
\label{fig:all_arch_test}
\end{figure}

When we test a trained Simease encoder and trajectory decoder (Fig. \ref{fig:all_arch_test}), we fix their neural network weights. Since we do not start with an entire trajectory, there is no interaction segment that provides an affordance embedding. Instead, the trained weights in Trajectory Encoder already obtained affordance knowledge due to the coupled triplet loss (Equ. \ref{eq:c_trip_loss}). Once the trajectory decoder outputs a predicted action $\hat{a}_t$ at step $t$, the $\hat{a}_t$, rather than a ground-truth action as in Equ. \ref{eq:e2}, would be used for the computation at step $t+1$.

\section{EVALUATION}

We design our experiments in order to answer the following questions: 1) How helpful is the coupled triplet loss? 2) How helpful is the contrastive learning framework with a deep Siamese network? 3) How good is our model in comparison with a state-of-the-art baseline?

To answer 1, we compare our full model with a version that uses a normal triplet loss; to answer 2, we compare our full model with a version that does not have Siamese network and is trained totally based on behavior cloning losses; to answer 3, we compare our full model with a recently published work \cite{morton2017simultaneous} as a baseline. Since we essentially solve an affordance-aware imitation learning problem for robotic grasping tasks, our evaluation metric involves grasping success rates which has been widely used for evaluating the learning of grasping/manipulation tasks (e.g. \cite{kleeberger2020survey,ren2020generalization}). 
We also show and analyze predicted affordance cues in a video from our project website$^1$ and partially in Fig. \ref{fig:example}.


\subsection{Evaluation Domain and Data Collection}

In our evaluation domain as shown in the leftmost region of Fig. \ref{fig:example}, we have a mug that is put in front of a Franka Panda Arm in the  PyBullet simulator \cite{coumans2016pybullet}. 
We use 24 mugs in our experiments. These mugs have different affordance characteristics and each belongs to one or more of the three affordance categories. The task for the robot is to pick up the mug and lift it up for 5 centimeters. To do this, the robot needs to intelligently infer the best way of grasping the mug, e.g., by its body, the left and right sides of its handle, or the front and back sides of its handle. This domain reflects a common service that humans may need from robots in our everyday lives. This domain is also sufficient for evaluating our algorithm due to the fact that naturally there are a large number of mugs that have different structures and geometric characteristics. It could even be necessary to consider totally different ways of grasping them from different locations.

When we collect data, we randomly select a mug model and put it in a predefined position. We fix its initial pose across all of our experiments to guarantee the existence of an affordable grasp that can be categorized into either of our three affordance categories. 
All demonstration trajectories are 8 steps long. We use PyBullet's function of reading users' debugging commands to interactively move the gripper to a good target pose with a mouse. Once the gripper is moved to an ideal target pose, the robot then executes with that target pose, and relevant information like observation, action, and state are recorded. We collect 27 trajectories for our training dataset. After every 10 training epochs, we test a model by randomly sampling 20 mugs and perform 20 grasps, respectively, and then record the success rate.

\subsection{Experimental Results and Analysis}
The quantitative performance is measured by grasping success rates and the qualitative performance is evaluated by showing predicted affordance cues in our video as mentioned before. The grasping results of our model, the baseline, and ablation study models are reported in Table \ref{tbl_result}. The success rate values are the highest testing success rates that a model achieves across all learning epochs. Table \ref{tbl_result} provides experiment results of grasping success rates for four models: 1. our Siamese encoder with coupled triplet loss; 2. our Siamese network without coupled triplet loss (we apply normal triplet loss on observation embeddings); 3. our model that is not trained in a contrastive learning framework, and 4. a baseline behavior cloning work \cite{morton2017simultaneous}.

\begin{table}[h]
\begin{center} {\footnotesize
\begin{tabular}{cc}
\hline
 Model 
 & Success Rate\\
\hline
1. Ours (Full Model): Siamese + Coupled Triplet Loss & 65\% \\[0ex]
2. Ours (Ablation): Siamese + Normal Triplet Loss & 35\% \\[0ex]
3. Ours (Ablation): Without Contrastive Learning & 45\% \\[0ex]
4. Baseline \cite{morton2017simultaneous} & 25\%\\[0ex] 
\hline
\end{tabular} }
\end{center}
\caption{}
\label{tbl_result}
\end{table}

The results clearly show the advantage of using our coupled triplet loss with a Siamese neural network for the learning of affordance cues and grasping from demonstrations. 
An interesting finding is that the model 3 still performs better than model 2. This suggests that merely doing contrastive learning at observation level via the normal triplet loss could misguide the policy learning. Instead, the coupled triplet loss in this work contrastively learn affordance embeddings to guide the learning of observation encoding. 

\section{CONCLUSIONS AND FUTURE WORKS}

We present an imitation learning algorithm that seeks training guidance not only from teachers' actions but from simultaneously discovering teachers' hidden affordance bias as well. We propose a contrastive learning framework with a Siamese encoder for affordance discovery and a trajectory decoder for policy learning. We represent affordance cues as visual attention. We further propose the coupled triplet loss to encourage the learned discriminative features to be more affordance-relevant. To the best of our knowledge, we initialize the direction of bridging the gap between affordance discovery and policy engineering/learning by achieving the two objectives together via an end-to-end deep neural network. Our work inherits the benefits of the class of works that learns affordances from demonstrations: there is no need of collecting ground-truth affordance labels for each image or pixel. However, such works focus on affordance learning and predicted affordances is still used by external motion planners. Our evaluation shows that our algorithm achieves the highest grasping success rate and predicts meaningful affordance cues. We believe our learning framework has a larger potential for solving more complex tasks, which could be pursued in the future extensions of this work:

1) The tasks that involve multiple levels of interactions, instead of only the interaction between gripper and mug. Most tool-using tasks would require multiple levels of interactions;  

2) The prediction of affordance-cue could be conditioned on different high-level tasks such that the attention map would be different even on the same object but w.r.t different tasks. In our work, we have a fixed high-level task of picking-to-pour-water, but if there is another task like pick-and-relocate, the discovered affordance cues might be different. This can be achieved by integrating our work into a hierarchical imitation learning framework.

3) It might also be interesting to explore how the coupled triplet loss could also be used to address other robot learning problems other than affordance-aware policy imitation.

\bibliographystyle{IEEEtranS.bst}
\bibliography{IEEEabrv.bib}

\end{document}